\documentclass[11pt,a4paper]{article}
\usepackage[utf8]{inputenc}
\usepackage[T1]{fontenc}
\usepackage{textcomp}
\usepackage[dvipsnames]{xcolor}
\usepackage[hyperref]{AACL-IJCNLP2020}
\usepackage{times}
\usepackage{latexsym}

\usepackage{url}
\usepackage{graphicx}
\usepackage{algorithm}
\usepackage{algpseudocode}
\usepackage{tabularx}
\usepackage{amsfonts}
\usepackage{hhline}
\usepackage{multirow}

\usepackage{microtype}
\usepackage{amsmath}

\usepackage{enumitem}

\aclfinalcopy 

\title{Systematically Exploring Redundancy Reduction in \\ Summarizing Long Documents}

\author{Wen Xiao and Giuseppe Carenini\\
  Department of Computer Science \\
  University of British Columbia \\
  Vancouver, BC, Canada, V6T 1Z4 \\
  {\tt \{xiaowen3, carenini\}@cs.ubc.ca}}

\date{}

\begin{document}
\maketitle
\begin{abstract}
Our analysis of large
summarization datasets indicates that redundancy is a very serious problem when summarizing long documents. 
Yet, redundancy reduction 
has not been thoroughly investigated in neural summarization. 
In this work, we systematically explore and compare different ways to deal with redundancy when summarizing long documents. Specifically, we organize the existing methods into categories based on when and how the redundancy is considered. Then, in the context of these categories, we propose three additional methods balancing non-redundancy and importance in a general and flexible way.
In a series of experiments, we show that our proposed methods achieve the state-of-the-art with respect to ROUGE scores on two scientific paper datasets, Pubmed and arXiv, while reducing redundancy significantly. \footnote{Our code can be found here -  \url{http://www.cs.ubc.ca/cs-research/lci/research-groups/natural-language-processing/}}

\end{abstract}

\section{Introduction}

Summarization is the task of shortening a given document(s) while maintaining the most important information. In general, a good summarizer should generate a summary that is syntactically accurate, semantically correct, coherent, and non-redundant  \cite{summary_eval}. While extractive methods tend to have better performance on the first two aspects, they are typically  less coherent and {\it more redundant} than abstractive ones, where new sentences are often generated by sentence fusion and compression, which helps detecting and removing redundancy \cite{abstractive_analysing}. Although eliminating redundancy has been  initially and more intensely studied  in the field of multi-document summarization \cite{Tackling_redundancy}, 
because important sentences selected from multiple documents (about the same topic) are more likely to be redundant than sentences from the same document, generating a non-redundant summary should still be one of the goals for single document summarization \cite{Lin2009}.

Generally speaking, there is a trade-off between importance and diversity (non-redundancy) \cite{Jung2019}, which is reflected in the two phases, \textit{sentence scoring} and \textit{sentence selection} \cite{NeuSum} in which extractive summarization task can be naturally decomposed. The former typically scores sentences based on importance, while the latter selects sentences based on their scores, but also possibly taking other factors (including redundancy) into account.

Traditionally, in non-neural approaches the trade-off between importance and redundancy has been carefully considered, with sentence selection picking sentences by optimizing an objective function that balances the two aspects \cite{mmr,redundancy_aware}. In contrast, more recent works on neural extractive summarization models has so far over-emphasized sentence importance and the corresponding scoring phase, while paying little attention to how to reduce redundancy in the selection phase, where they simply apply a greedy algorithm to select sentences (e.g.,\citet{cheng-lapata-2016-neural, xiao-carenini-2019-extractive}). Notice that this is especially problematic for long documents, where redundancy tends to be a more serious problem, as we have observed in key datasets. Improving redundancy reduction in neural extractive summarization for long documents is a major goal of this paper.

Indeed, some recently proposed neural methods
aim to reduce redundancy, but they either do that implicitly or inflexibly and only focusing on short documents (e.g., news). For instance, some  models learn to reduce redundancy when predicting the scores \cite{summarunner}, or jointly learn to score and select sentences \cite{NeuSum} in an implicit way. However, whether these strategies actually help reducing redundancy 
is still an open empirical question. The only neural attempt of explicitly reduce redundancy in the sentence selection phase is the Trigram Blocking technique, used in  recent extractive summarization models on news datasets (e.g., \cite{liu-lapata-2019-text}). However, the effectiveness of such strategy on the summarization of long documents has not been tested. Finally, a very recent work by \citet{Bi2020} attempts to reduce redundancy in more sophisticated ways, but still focusing on news. Furthermore, since it relies on BERT, such model is unsuitable to deal with long documents (with over 3,000 words).



To address this rather confusing situation, characterized by unclear connections between all the proposed neural models, by their limited focus on short documents, and by spotty evaluations,
in this paper we systematically organize existing redundancy reduction methods into three categories, and compare them with respect to the informativeness and redundancy of the generated summary for long documents. In particular, to perform a fair comparison we re-implement all methods by modifying a common basic model \cite{xiao-carenini-2019-extractive}, which is a top performer on long documents without considering redundancy.
Additionally,  we propose three new methods that we argue will reduce  redundancy more explicitly and flexibly in the sentence scoring and sentence selection phase by deploying more suitable decoders, loss functions and/or sentence selection algorithms, again building for a fair comparison on the  common basic model \cite{xiao-carenini-2019-extractive}.

To summarize, our main contributions in this paper are: we first examine popular datasets, and show that redundancy is a more serious problem when summarizing long documents (e.g., scientific papers) than short ones (e.g. news). Secondly, we not only  reorganize and re-implement existing neural methods for redundancy reduction, but we also propose three new general and  flexible methods. 
Finally, in a series of experiments, we compare existing and proposed methods 
on long documents (i.e., the Pubmed and arXiv datasets), with respect to ROUGE scores \cite{lin-2004-rouge} and redundancy scores \cite{peyrard-etal-2017-learning, nid_score}.

As a preview, empirical results reveal that the proposed methods achieve state-of-the-art performance on ROUGE scores, on the two scientific paper datasets, while also reducing the redundancy significantly. 

\section{Related Work}



In traditional extractive summarization, 
the process is treated as a discrete optimization problem balancing between importance scores and redundancy scores, with techniques like Maximal Marginal Relevance(MMR)\cite{mmr}, redundancy-aware feature-based sentence classifiers \cite{redundancy_aware} and graph-based submodular selection \cite{Lin2009}. 

In recent years, researchers have explored neural extractive summarization solutions, which score sentences by training the neural models on a large corpus, and simply apply a greedy algorithm for sentence selection \cite{cheng-lapata-2016-neural,summarunner}. Although a model with a sequence decoder might plausibly encode  redundancy information implicitly, \newcite{kedzie-etal-2018-content} empirically show that this is not the case, since non auto-regressive models (the ones scoring each sentence independently), perform on par with 
models with a sequence decoder. In one of our new methods, 
to effectively capture redundancy information, we specify a new loss that explicitly consider redundancy when training the neural model.

Beyond a greedy algorithm, the Trigram Blocking is frequently used to explicitly reduce redundancy in the sentence selection phase 
\cite{liu-lapata-2019-text}. In essence, a new sentence is not added to the summary if it shares a 3-gram with the previously added one.
\newcite{rl-abstractive} first adopt the strategy for abstractive summarization, which forces the model not to produce the same trigram twice in the generated summaries, as a simplified version of MMR \cite{mmr}.  Arguably, this method is too crude for documents with relatively long sentences or specific concentrations (e.g. scientific papers), where some technical terms, possibly longer than 2-grams, are repeated frequently in the 'important sentences' (even in the reference summaries).
To address this limitation, we propose a neural version of MMR to deal with redundancy within the sentence selection phase in a more flexible way, that can be tuned to balance importance and non-redundancy as needed.

The idea of MMR has also inspired 
\newcite{NeuSum}, who propose a model jointly learning to score and select the sentences. Yet, this work not only focuses on summarizing short documents (i.e., news), but also uses MMR implicitly, and arguably sub-optimally, by learning a score that only indirectly captures the trade-off between relevance and redundancy. To improve on this approach, in this paper we propose 
a third new method, in which importance and redundancy are explicitly weighted, while still making the sentence scoring and selection benefit from each other by fine tuning the trained neural model through a Reinforcement Learning (RL) mechanism.

Finally, \citet{Bi2020} is the most recent (still unpublished) work on reducing redundancy in neural single document summarization. However, their goal is very different form ours, since they focus on relatively short documents in the news domain.


\section{Measuring Redundancy: metrics and comparing long vs. short documents}
\label{sec-measure-red}
We use the following two relatively new metrics to measure redundancy in the source documents and in the generated summaries.

\textbf{Unique n-gram ratio\footnote{In this paper, all the unique n-gram ratios are shown in percentage.}:} proposed in \newcite{peyrard-etal-2017-learning}, it measures n-grams uniqueness; the lower it is, the more redundant the document is.  $$Uniq\_ngram\_ratio = \frac{count(uniq\_n\_gram)}{count(n\_gram)}$$

\textbf{Normalized Inverse of Diversity (NID):} captures redundancy, as the inverse of a diversity metric with length normalization. Diversity is defined as the entropy of unigrams in the document \cite{nid_score}.
Since longer documents are more likely to have a higher entropy, we normalize the diversity with the maximum possible entropy for the document $log(|D|)$. Thus, we have:
\begin{align*}
  NID &= 1-\frac{entropy(D)}{log(|D|)}
    \end{align*}
Note that higher NID indicates more redundancy. \\

\begin{figure}
    \centering
    \includegraphics[width=0.9\linewidth]{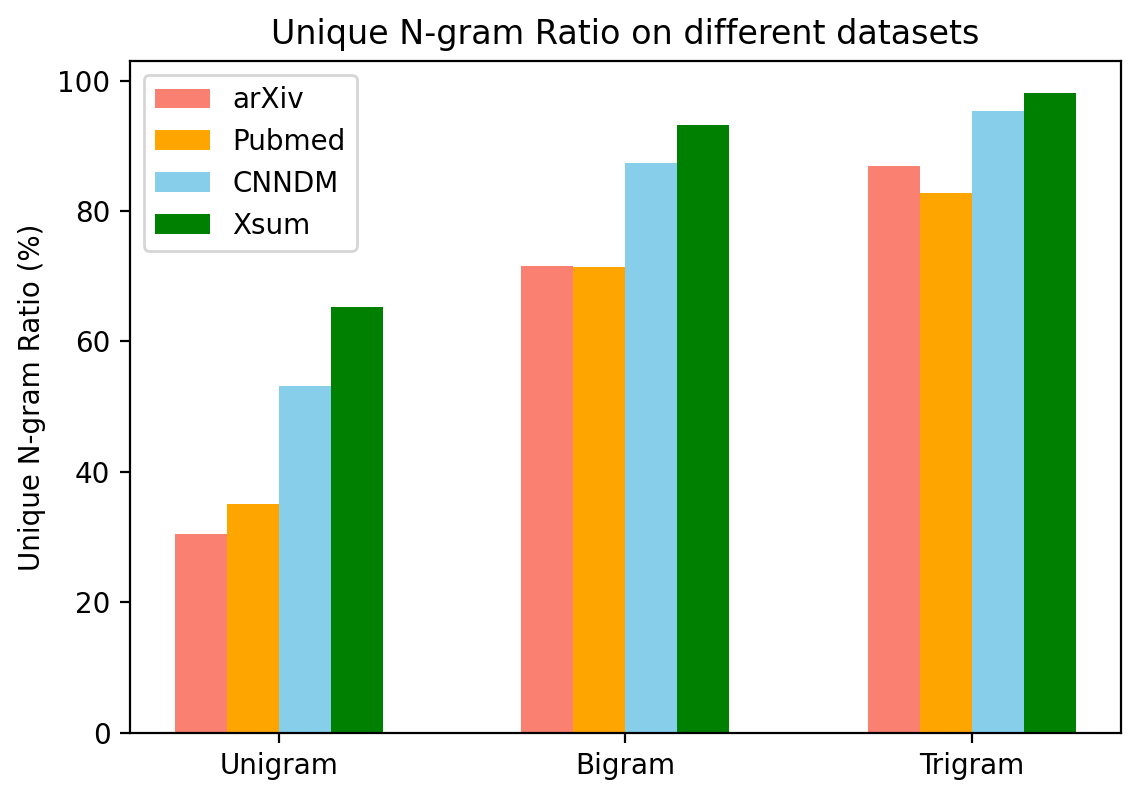}
    \caption{The average unique n-gram ratio in the documents across different datasets. To reduce the effect of length difference, stopwords were removed.}
    \label{fig:n-gram}
    \vspace{-4mm}
\end{figure}

When we compare the redundancy of long vs. short documents with respect to these two metrics  on four popular datasets for summarization (CNNDM \cite{nallapati-etal-2016-abstractive}, Xsum \cite{narayan-etal-2018-dont}, Pubmed and arXiv \cite{discourse-aware}), we observe that long documents are substantially more redundant than short ones (as it was already pointed out in the past \cite{longdoc_more_redundancy}).    
Table \ref{tab:inverse-diversity} shows the basic statistics of each dataset, along with the average NID scores, while
Figure \ref{fig:n-gram} shows the average Unique n-gram Ratio for the same datasets. These observations provide further evidence that redundancy is a more serious problem in long documents. In addition, 
notice that the sentences in the scientific paper datasets are much longer than in the news datasets, which plausibly makes it even harder to balance between importance and non-redundancy.

\begin{table}[h!]
    \centering
    \resizebox{\linewidth}{!}{
    \begin{tabular}{c|c|c|c|c}
    \hline
       Datasets  &\# Doc. & \# words/doc. & \# words/sent.& NID \\
       \hline
       Xsum & 203k &429 &22.8 & 0.188\\
       \hline
       CNNDM &270k &  823&19.9 &0.205\\
       \hline
       
       Pubmed&115k &3142 &35.1 &0.255\\
       \hline
       arXiv &201k &6081&29.2 &0.267\\
       \hline
    \end{tabular}}
    \caption{Longer documents are more redundant 
    }
    \label{tab:inverse-diversity}
    \vspace{-5mm}
\end{table}

\section{Redundancy Reduction Methods}
\begin{table*}[]
    \centering
    \resizebox{0.9\linewidth}{!}{
    \begin{tabular}{c|l|c|c|c|c}
    
\multirow{2}{*}{\textit{Categ.}}&\multirow{2}{*}{\textit{Methods}}& \multicolumn{3}{c|}{\textit{Sent. Scor.}} &\multirow{2}{*}{\textit{Sent. Sel.}}\\
\cline{3-5}
& & \textit{Encoder}&\textit{Decoder}&\textit{Loss Func.}\\
\hhline{======}
-&Naive MMR& \multicolumn{3}{c|}{ Cosine Similarity}& MMR Select\\
\hline
-&ExtSum-LG&Encoder-LG&MLP&Cross Entropy (CE)&Greedy\\
\hline
A&\quad+ SR Decoder&Encoder-LG&SR Decoder&CE&Greedy\\
\hline

A&\quad+ NeuSum Decoder&Encoder-LG&NeuSum Decoder&KL Divergence&NeuSum Decoder\\
\hline
B&\quad\textbf{+ RdLoss}&Encoder-LG&MLP&CE + 
Red. Loss1
&Greedy\\
\hline
C&\quad+ Trigram Blocking&Encoder-LG&MLP&CE&Trigram Blocking\\
\hline
C&\quad\textbf{+ MMR-Select}&Encoder-LG&MLP&CE&MMR Select\\
\hline
C&\quad\textbf{+ MMR-Select+}&Encoder-LG&MLP&CE + Red. Loss2&MMR Select\\
\hline

    \end{tabular}}
    \caption{The architecture of redundancy reduction methods. \textbf{Bold} methods are proposed in this paper.}
    \label{tab:model_to_compare}
\end{table*}

We systematically organize neural redundancy reduction methods into three categories, and compare prototypical methods from each category.
\vspace{-1mm}
\renewcommand{\labelenumi}{\Alph{enumi}}
\begin{enumerate}

    \item The decoder is designed to implicitly take redundancy into account.
\item In the sentence scoring phase, explicitly learn to reduce the redundancy.
\item In the sentence selection phase, select sentences with less redundancy.
\end{enumerate}
\vspace{-1mm}

In this section, we describe different methods from each category. To compare them in a fair way, we build all of them on a basic ExtSum-LG model (see  \S\ref{basic-model}), by modifying the decoder and the loss function in the sentence selection phase or the sentence selection algorithm. In Table \ref{tab:model_to_compare}, we summarize the architecture (Encoder, Decoder, Loss Function and sentence selection algorithm) of all the methods we compare.

\subsection{Baseline Models}
\label{basic-model}
We consider two baseline models. One is an influential unsupervised method explicitly balancing importance and  redundancy (Naive MMR). The other is our basic neural supervised model not dealing with redundancy at all (ExtSum-LG), to which we add different redundancy reduction mechanisms.

\subsubsection*{Naive MMR}
MMR \cite{mmr} is a traditional extractive summarization method, which re-ranks the candidate sentences with a balance between query-relevance(importance) and information novelty(non-redundancy).
Given a document $D$, at each step, MMR selects one sentence from the candidate set $D \setminus \hat{S}$ that is relevant with the query $Q$, while containing little redundancy with the current summary $\hat{S}$. Note that if there is no specific query, then the query is the representation of the whole document. The method can be formally specified as:
\vspace{-2mm}
\begin{align*}
MMR &= \arg\max_{s_i \in D \setminus \hat{S}}[\lambda Sim_1(s_i,Q)\\
&\quad-(1-\lambda)\max_{s_j\in \hat{S}}Sim_2(s_i,s_j)]
\end{align*}
where $Sim_1(s_i,Q)$ measures the similarity between the candidate sentence $s_i$ and the query, indicating the importance of $s_i$, while  $\max_{s_j\in \hat{S}}Sim_2(s_i,s_j)$ measures the similarity between the candidate sentence $s_i$ and the current summary $\hat{S}$, representing the redundancy, and $\lambda$ is the balancing factor. In this work, all the $Sim$ are computed as the cosine similarity between 
the embeddings of the sentences. 

\subsubsection*{ExtSum-LG}
For the basic model, we use the current state-of-the-art model \cite{xiao-carenini-2019-extractive} on the summarization of long documents. It is a novel extractive summarization model incorporating local context and global context in the encoder, with an MLP layer as decoder and cross-entropy as the loss function. For the sentence selection phase, it greedily picks the sentences according to the score predicted by the neural model. In this method, redundancy is not considered, so it is a good testbed for adding and comparing redundancy reduction methods.

Specifically, for a document $D=\{s_1,s_2,...,s_n\}$, the output of the encoder is $h_i$ for each sentence $s_i$, and the decoder gives output $P(y_i)$ as the confidence score on the importance of sentence $s_i$. Finally, the model is trained on the Cross Entropy Loss :
\vspace{-1mm}
$$L_{ce} = -\sum_{i=1}^{n}(y_i\log{P(y_i)}+(1-y_i)\log{(1-P(y_i))}$$
\subsection{Implicitly Reduce Redundancy in the neural model (Category A, Table \ref{tab:model_to_compare})}
In this section, we describe two decoders from previous work, in which the redundancy of the summary is considered implicitly.

\textbf{SummaRuNNer Decoder:}
\newcite{summarunner} introduce
a decoder that computes a sentence score based on its salience, novelty(non-redundancy) and position to decide whether it should be included in the 
summary.  
Formally: 
\vspace{-1mm}
\begin{eqnarray*}
P(y_i) = \sigma(W_ch_i & & \# Content\\
        +h_iW_sd& & \# Salience\\
        -h_i^TW_r\tanh(summ_i) & & \# Novelty\\
        +W_{ap}p_i^a+W_{rp}p_i^r & &\# Position\\
        +b) & & \# Bias
\end{eqnarray*}
\vspace{-1mm}
where $h_i$ is the hidden state of sentence $i$ from the encoder, $d$ is the document representation 
, $summ_i$ is the summary representation, updated after each decoding step 
, and $p_i^a$, $p_i^r$ are absolute and relative position embeddings, respectively.
Once $P(y_i)$ is obtained for each sentence $i$, a greedy algorithm 
selects the sentences to form the final summary. Notice that although SummaRuNNer does contain a component assessing novelty, it would be inappropriate to view this model as explicitly dealing with redunadany because   the novelty component is not directly supervised.

\textbf{NeuSum Decoder:} 
One of the main drawback of SummaRuNNer decoder is that it always score the sentences in order, i.e., the former sentences are not influenced by the latter ones. In addition, it only considers redundancy in the sentence scoring phase, while simply using a greedy algorithm to select sentences according to the resulting scores. To address these problems, \newcite{NeuSum} propose a new decoder to identify the relative gain of sentences, jointly learning to score and select sentences. In such decoder, instead of feeding the sentences and getting the scores in order, they use a mechanism similar to the pointer network \cite{pointer_network} to 
predict the scores of all the sentences at each step, select the sentence with the highest score, and feed it to the next step of sentence selection. As for the loss function, they use the KL divergence between the predicted score distribution and the relative ROUGE F1 gain at each step. 
To be specific, the loss computed at step $t$ is: 
\vspace{-2mm}
\begin{eqnarray*}
L_t &=& D_{KL}(P_t||Q_t)\\
P_t(y_i) &=&\frac{\exp(\sigma(h_i))}{\sum_{j=1}^n{\exp(\sigma(h_j))}}\\
Q_t(y_i) &=& \frac{\exp(\tau \Tilde{g}_t(y_i))}{\sum_{j=1}^n{\exp(\tau \Tilde{g}_t(y_j))}}\\
g_t(y_i) &=& r1(\mathbb{S}_{t-1}\cup s_i)-r1(\mathbb{S}_{t-1})
\end{eqnarray*}
where $P_t$, $Q_t$ are the predicted and ground truth relative gain respectively, $g_t(y_i)$ is the ROUGE F1 gain with respect to the current partial summary $\mathbb{S}_{t-1}$ for sentence $s_i$, and $\Tilde{g}_t(y_i)$ is the Min-Max normalized $g_t(y_i)$. $\tau$ is a smoothing factor, which is set to $200$ empirically on the Pubmed dataset. \footnote{Due to the complexity of generating the target distribution $Q$, we only experiment with this method on Pubmed.}

\begin{figure*}[h!]
    \centering
    \includegraphics[width=0.70\linewidth]{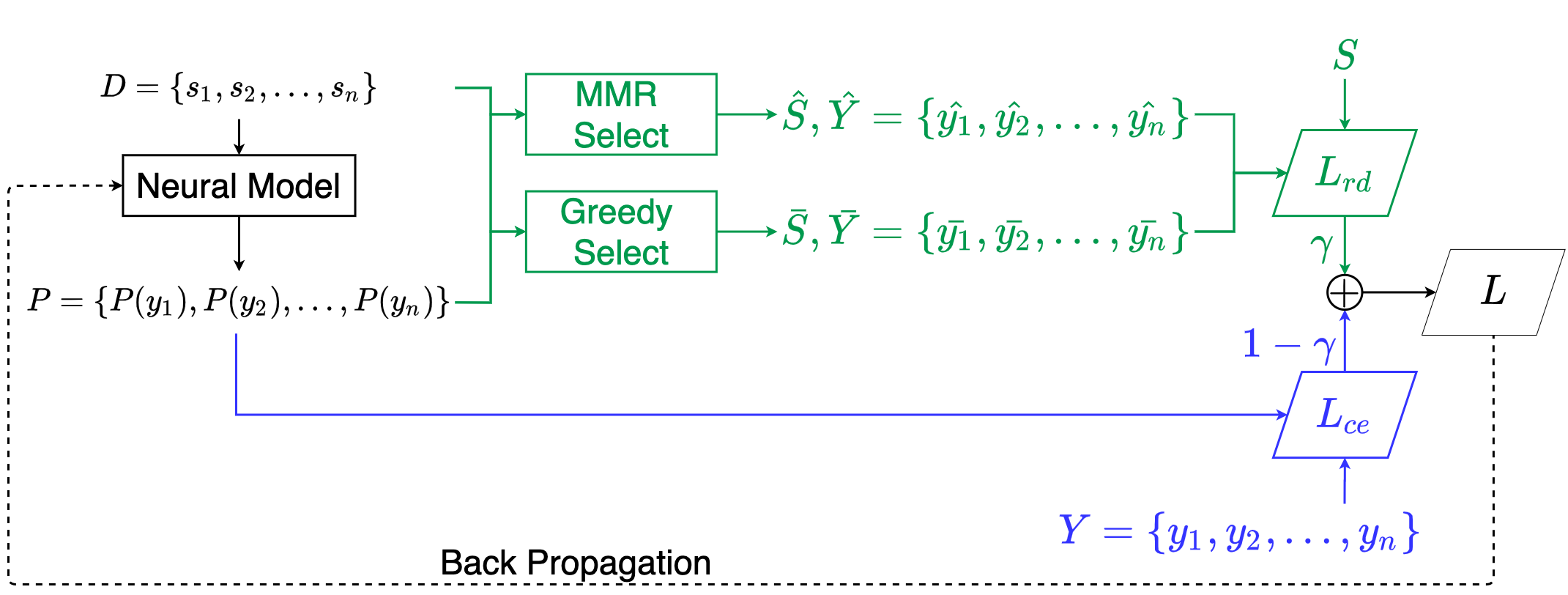}
    \caption{The pipeline of the MMR-Select+ method, where $\hat{S},\hat{Y}$ and $\bar{S},\bar{Y}$  are the summary and labels generated by the MMR-Select algorithm and the normal greedy algorithm, respectively. $S$ and $Y$ are the ground truth summary and the oracle labels.}
    \label{fig:mmr-rl}
    \vspace{-2mm}
\end{figure*}
\vspace{-1mm}
\subsection{Explicitly Reduce Redundancy in Sentence Scoring (Category B, Table \ref{tab:model_to_compare})}
We propose a new method to explicitly learn to reduce redundancy when scoring the sentences.

\textbf{RdLoss:}
Although \newcite{NeuSum} jointly train the decoder to score and select sentences, it still learns to reduce redundancy implicitly, and the method does not allow controlling the degree of redundancy. To address this limitation, we propose a rather simple method to explicitly force the model to reduce redundancy in the sentence scoring phase by adding 
a redundancy loss term
to the original loss function, motivated by the success of a similar strategy of adding a bias loss term in the gender debiasing task 
\cite{qian-etal-2019-reducing}. Our new loss term $L_{rd}$ is naturally defined as the expected redundancy contained in the resulting summary, as shown below:
\vspace{-2mm}
\begin{eqnarray*}
L &=& \beta L_{ce}+ (1-\beta) L_{rd}\\
L_{rd}&=&\sum_{i=1}^{n} \sum_{j=1}^{n} P(y_i)P(y_j)Sim(s_i, s_j)
\end{eqnarray*}
where $P(y_i),P(y_j)$ are the confidence scores of sentence $i$ and $j$ on whether to select the sentences in the generated summary, and $Sim(s_i,s_j)$ is the similarity, i.e. redundancy between sentence $i$ and $j$. \footnote{Noting that we define $Sim(s_i,s_i)$ as $0$} By adding the redundancy loss term, we penalize it more if two sentences are similar to each other and both of them have high confidence scores. $\beta$ is a balance factor, controlling the degree of redundancy.
\vspace{-1mm}
\subsection{Explicitly Reduce Redundancy in Sentence Selection (Category C, Table \ref{tab:model_to_compare})}
We first introduce an existing method and then propose two novel methods that explicitly reduce redundancy in the sentence selection phase.

\textbf{Trigram Blocking} is widely used in  recent extractive summarization models on the news
dataset (e.g. \citet{liu-lapata-2019-text}). Intuitively, it borrows the idea of MMR to balance the importance and non-redundancy when selecting sentences. In particular, given the predicted sentence scores, instead of just selecting sentences greedily according to the scores, 
the current candidate is added to the summary 
only if it does not have trigram overlap with the previous selected sentences. Otherwise, the current candidate sentence is ignored and the next one is checked, until the length limit is reached.

\textbf{MMR-Select:}
Inspired by the existence of a relevance/redundancy trade-off, we propose MMR-Select, a simple method to eliminate redundancy when a neural summarizer selects sentences to form a summary, in a way that is arguably more flexible than Trigram Blocking with a balance factor $\lambda$.
  
 With the confidence score computed by the basic model, $P=\{P(y_1),P(y_2),...,P(y_n)\}$, instead of picking sentences greedily, we pick the sentences according to the MMR-score, which is defined based on MMR and updated after each single sentence being selected.
\vspace{-2mm}
 \begin{align*}
 \text{MMR-Select} &= \arg\max_{s_i \in D \setminus \hat{S}}[\text{MMR-score}_i]\\
\text{MMR-score}_i &= \lambda P(y_i) -(1-\lambda)\max_{s_j\in \hat{S}}Sim(s_i,s_j)]
\end{align*}
The main difference between the Naive MMR and MMR-Select falls into the computation of the importance score. In the Naive MMR, the importance score is the similarity between each sentence and the query, or the whole document, while in MMR-Select, the importance score is computed by a trained neural model. 

\textbf{MMR-Select+ :}
The main limitation of MMR-Select is that the sentence scoring phase and the sentence selection phase cannot benefit from each other, because they are totally separate.

To promote synergy between these two phases, we design a new method, MMR-Select+, shown in Figure \ref{fig:mmr-rl}, which synergistically combines three components: the basic model, the original cross-entropy loss $L_{ce}$(in blue), and an RL mechanism (in green) whose loss is $L_{rd}$. The neural model is then trained on a mixed objective loss $L$ with $\gamma$ as the scaling factor. Zooming on the details of the RL component, it first generates a summary $\hat{S}$ by applying the  MMR selection described for MMR-Select, which is to greedily pick sentences according to MMR-score, as well as the corresponding label assignment $\hat{Y}=\{\hat{y_1},\hat{y_2},...,\hat{y_n}\}$ ($\hat{y_i}=1$ if $s_i$ is selected, $\hat{y_i}=0$ otherwise). Then, the expected reward is computed based on the ROUGE score between $\hat{S}$ and the gold-standard human abstractive summary $S$ weighted by the probability of the $\hat{Y}$ labels. Notice that we also adopt the self-critical strategy \cite{rl-abstractive} to help accelerating the convergence by adding a baseline summary$\bar{S}$, which is generated by greedily picking the sentences according to $P$. $r(\bar{S})$ is the reward of this baseline summary and it is subtracted from $r(\hat{S})$ to only positively reward summaries which are better than the baseline. Formally, the whole MMR-Select+ model can be specified as follows:
\vspace{-2mm}
\begin{align*}
    L &= \gamma L_{rd}+(1-\gamma) L_{ce} \\
    L_{rd}  
     &= -(r(\hat{S})-r(\bar{S}))\sum_{i=1}^{n}\log P(\hat{y_i}) \\
     r(S') &=  \frac{1}{3}\sum_{k\in \{1,2,L\}}\text{ROUGE-k}(S',S)
\end{align*}
\vspace{-2mm}
\begin{figure*}
    \centering
    \includegraphics[width=0.9\linewidth]{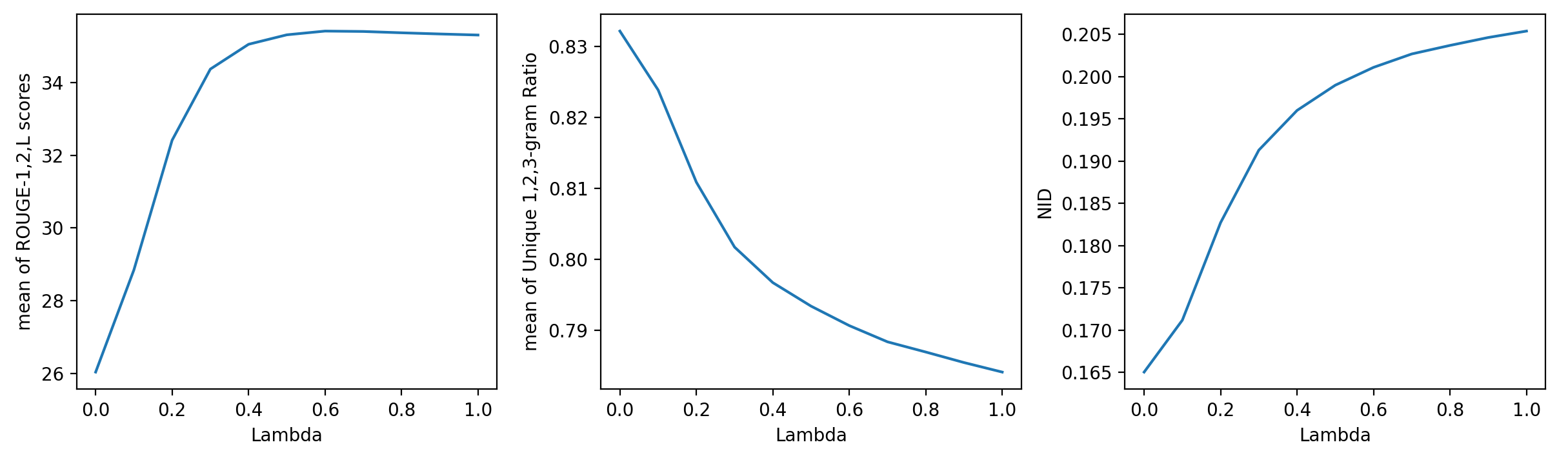}
    \vspace{-1mm}
    \caption{The average ROUGE scores, average unique n-gram ratios, and average NID scores with different $\lambda$ used in the MMR-Select on the validation set. Remember that the higher the Unique n-gram Ratio, the lower NID, the less redundancy contained in the summary.}
    \label{fig:finetune}
\end{figure*}

 \begin{table*}[h!]
    \centering
    \resizebox{0.85\linewidth}{!}{
    \begin{tabular}{c|l|c|c|c|c|c|c}
    \hline
    \multirow{2}{*}{Categ.}&\multirow{2}{*}{Model}&\multicolumn{3}{c|}{Pubmed}&\multicolumn{3}{c}{arXiv}\\
    \cline{3-8}
      &   &ROUGE-1&ROUGE-2&ROUGE-L &ROUGE-1&ROUGE-2&ROUGE-L \\
    \hline
C&       Naive MMR&37.46 &11.25 &32.22&33.74 &8.50 &28.36\\
-&       ExtSum-LG\footnotemark  & 45.18 & 20.20&40.72& 43.77 & 17.50&38.71\\
\hline
A&       \quad+SR Decoder &45.18 & \textcolor{Red}{20.16}&\textcolor{Red}{40.69}&\textcolor{Green}{43.92} &\textcolor{Green}{17.65} &\textcolor{Green}{38.83}\\
A&       \quad+NeuSum Decoder &\textcolor{Red}{44.54}&\textcolor{Red}{19.66}&\textcolor{Red}{40.42}&- &- &-\\
B&        \quad\textbf{+RdLoss}&{\color{Green}45.30} $\dagger$ &\textcolor{Green}{\textbf{20.42}} $\dagger$ &\textcolor{Green}{40.95} $\dagger$ &\textcolor{Green}{\textbf{44.01}} $\dagger$ &\textcolor{Green}{\textbf{17.79}} $\dagger$ &\textcolor{Green}{\textbf{39.09}} $\dagger$\\       
C&       \quad+Trigram Blocking &\textcolor{Red}{43.33}&\textcolor{Red}{17.67}&\textcolor{Red}{39.01}&\textcolor{Red}{42.75} &\textcolor{Red}{15.73} &\textcolor{Red}{37.85}\\
C&       \quad\textbf{+MMR-Select}& \textcolor{Green}{45.29} $\dagger$& \textcolor{Green}{20.30} $\dagger$ &\textcolor{Green}{40.90} $\dagger$ &\textcolor{Green}{43.81} &\textcolor{Red}{17.41}&\textcolor{Green}{38.94}\\
C&        \quad\textbf{+MMR-Select+} &\textcolor{Green}{\textbf{45.39}} $\dagger$ &\textcolor{Green}{20.37} $\dagger$ &\textcolor{Green}{\textbf{40.99}} $\dagger$&\textcolor{Green}{43.87} $\dagger$ &17.50&\textcolor{Green}{38.97} $\dagger$\\

    \hline
-&       Oracle &55.05 &27.48 &49.11&53.89 &23.07 &46.54\\
       \hline
    \end{tabular}}
    \caption{Rouge score of different summarization models on the Pubmed and arXiv datasets. $\dagger$ indicates significantly better than the ExtSum-LG with confidence level 99\% on the Bootstrap Significance test. \textcolor{Green}{Green} numbers means it's \textbf{better} than ExtSum-LG on the certain metric, and the \textcolor{Red}{red} numbers means worse.}
    \label{tab:rouge}
\end{table*}

\begin{table*}[h!]
    \centering
    \resizebox{0.9\linewidth}{!}{
    \begin{tabular}{c|l|c|c|c|c|c|c|c|c}
    \hline
    \multirow{2}{*}{Categ.}&\multirow{2}{*}{Model}&\multicolumn{4}{c|}{Pubmed}&\multicolumn{4}{c}{arXiv}\\
    \cline{3-10}
  &   & Unigram\% & Bigram\%& Trigram\%& NID& Unigram\% & Bigram\%& Trigram\%& NID\\
    \hline
C& Naive MMR&56.55& 90.93&96.95 &0.1881&53.01&88.82 &96.28&0.1992\\
-& ExtSum-LG  &53.02  &87.29&94.37 &0.2066&52.17  &87.19&95.38 &0.2088\\
\hline
A&  \quad+SR Decoder& \textcolor{Red}{52.88}&\textcolor{Red}{87.17}&\textcolor{Red}{94.32}&\textcolor{Red}{ 0.2070 }&\textcolor{Red}{51.98} &\textcolor{Red}{87.08} &\textcolor{Red}{95.31}&\textcolor{Red}{0.2097}\\
A&        \quad+NeuSum Decoder&\textcolor{Green}{54.88} $\dagger$&\textcolor{Green}{88.71} $\dagger$&\textcolor{Green}{95.13} $\dagger$&\textcolor{Green}{0.1993} $\dagger$&- &- &-&-\\
B&         \quad\textbf{+RdLoss}&\textcolor{Green}{53.23} $\dagger$ &\color{Green}87.41 &\color{Green}94.43 &\textcolor{Green}{0.2052} $\dagger$ &52.17&\color{Green}87.20 &\textcolor{Red}{95.36} &\color{Green}0.2085\\

C&        \quad+Trigram Blocking &\textcolor{Green}{\textbf{57.58}} $\dagger$ $\ddagger$&\textcolor{Green}{\textbf{93.05}} $\dagger$ $\ddagger$&\textcolor{Green}{\textbf{98.56}} $\dagger$ $\ddagger$&\textcolor{Green}{\textbf{0.1818}} $\dagger$ $\ddagger$&\textcolor{Green}{\textbf{56.12}} $\dagger$ $\ddagger$&\textcolor{Green}{\textbf{92.38}} $\dagger$ $\ddagger$ &\textcolor{Green}{\textbf{98.94}} $\dagger$ $\ddagger$ &\textcolor{Green}{\textbf{0.1876}} $\dagger$ $\ddagger$\\
C&       \quad\textbf{+MMR-Select}  & \textcolor{Green}{53.76} $\dagger$ &\textcolor{Green}{88.04} $\dagger$ & \textcolor{Green}{94.96} $\dagger$ &\textcolor{Green}{0.2022}&\textcolor{Green}{52.80} $\dagger$  &\textcolor{Green}{87.64} $\dagger$ &\color{Green}95.40&\textcolor{Green}{0.2055} $\dagger$\\
C&         \quad\textbf{+MMR-Select+} &\textcolor{Green}{53.93} $\dagger$ &\color{Green}88.32 &\color{Green}95.14 &\color{Green}0.2014&\textcolor{Green}{52.76} $\dagger$ &\textcolor{Green}{87.78} $\dagger$ &\textcolor{Green}{95.70} $\dagger$&\textcolor{Green}{0.2055} $\dagger$ \\
         
        \hline
-&        Oracle &56.66 &89.25 & 95.55&0.2036& 56.74 &90.81 & 96.82&0.2029\\
        \hline
-&        Reference&56.69 & 89.45&95.95 &0.2005&58.92 & 90.13& 97.02 &0.1970\\
        \hline
    \end{tabular}}
    \caption{Unique n-gram ratio and NID score on the two datasets. $\dagger$ indicates significant differences from \cite{xiao-carenini-2019-extractive} with confidence level 99\%, while $\ddagger$ indicates significant differences from all the other models with confidence level 99\% on the Bootstrap Significance test. Noting the higher the Unique n-gram Ratio, the lower NID, the less redundancy contained in the summary.\textcolor{Green}{Green} numbers means it's \textbf{better }than ExtSum-LG on the certain metric, and the \textcolor{Red}{red} numbers means worse.}
    \label{tab:redundancy}
    \vspace{-2mm}
\end{table*}
\vspace{-1mm}
\section{Experiments}
In this section, we describe the settings, results and analysis of the experiments of different methods on the Pubmed and arXiv datasets.
\vspace{-1mm}
\subsection{Model Settings}

\footnotetext{The results of ExtSum-LG were obtained by re-running their model. 
}

Following previous work, 
we use GloVe \cite{glove} as word embedding, and the average word embedding as the distributed representation of sentences. To be comparable with \citet{xiao-carenini-2019-extractive}, we set word length limit of the generated summaries as $200$ on both datasets.  \footnote{A document representation in Unsupervised MMR is similarly computed by averaging the embeddings of all the words.}
We tune the hyperparameter $\lambda$ and $\beta$ in the respective methods on the validation set, and set $\lambda=0.6, \beta=0.3$ for both datasets. Following previous work (e.g., \citet{bertrl}), $\gamma$ was set to $0.99$. For training MMR-Select+, the learning rate is $lr=1e-6$; we start with the pretrained ExtSumm-LG model. As for the evaluation metric, we use ROUGE scores as the measurement of importance while using the Unique N-gram Ratio and NID defined in Section \ref{sec-measure-red} as the measurements of redundancy.
\vspace{-1mm}
\subsection{Finetuning $\lambda$}
\vspace{-1mm}
Consistently with previous work \cite{Jung2019}, when we finetune $\lambda$ of MMR Select on the validation set, we pinpoint the trade off between importance and non-redundancy in the generated summary (see Figure \ref{fig:finetune}). For $\lambda\leq0.6$, as we increase the weight of importance score, the average ROUGE scores continuously increase while the redundancy/diversity increases/drops rapidly. But since extractive methods can only reuse sentences from the input document, there is an upper bound on how much the generated summary can match the ground-truth summary, so when $\lambda > 0.6$, 
the ROUGE score even drops by a small margin, while the redundancy/diversity still increases/drops.
Then the problem to solve for future work is how to increase the peak, which could be done by either applying finer units (e.g., clauses instead of sentences) or further improve the model that predicts the importance score.
\vspace{-1mm}
\subsection{Overall Results and Analysis}
\label{result_and_analysis}
\vspace{-1mm}
The experimental results for the ROUGE scores are shown in Table \ref{tab:rouge}, whereas results for redundancy scores (Unique N-gram Ratio and NID score) are shown in Table \ref{tab:redundancy}.
With respect to the balance between importance and non-redundancy, despite the trade-off between the two aspects, all of the three methods we propose can reduce redundancy significantly while also improving the ROUGE score significantly compared with the ExtSum-LG basic neural model. In contrast, the NeuSum Decoder and Trigram Blocking effectively reduce  redundancy, but in doing that they hurt the importance aspect considerably. Even worse, the SR Decoder is dominated by the basic model on both aspects.  


Focusing on the redundancy aspect (Table \ref{tab:redundancy}), Trigram Blocking makes the largest improvement on redundancy reduction, but with a large drop in ROUGE scores. This is in striking contrast with results on news datasets \cite{liu-lapata-2019-text}, where Trigram Blocking reduced redundancy while also improving the ROUGE score significantly. Plausibly,  the difference between the performances across datasets might be the result of the inflexibility of the method. In both Pubmed and arXiv datasets, the sentences are much longer than those in the news dataset (See Table \ref{tab:inverse-diversity}), and therefore, simply dropping candidate sentences with 3-gram overlap may lead to incorrectly missing sentences with substantial important information. 

Furthermore, another insight revealed in Table \ref{tab:redundancy} is that dealing with redundancy in the sentence selection phase is consistently more effective than doing it in the sentence scoring phase, regardless of whether this happens implicitly (NeuSum $>$ SR Decoder) or explicitly (Trigram Blocking, MMR-Select/+ $>$ RdLoss).

Moving to more specific findings about particular systems, we already noted that while the NeuSum Decoder reduces redundancy effectively, it performs poorly on the ROUGE score, something that did not happen with news datasets. A possible explanation is that 
the number of sentences selected for the scientific paper datasets (on average 6-7 sentences) is almost twice the number of sentences selected for news;  and as it was mentioned in the original paper \cite{NeuSum}, the precision of NeuSum drops rapidly after selecting a few sentences.

Other results confirm established beliefs. 
The considerable difference between Naive MMR and MMR-Select was expected given the recognized power of neural network over unsupervised methods. 
Secondly, the unimpressive performance of the SR decoder confirms that the in-order sequence scoring is too limited for effectively predicting importance score and reducing redundancy. 


\begin{figure*}[h!]
    \centering
    \includegraphics[width=0.8\linewidth]{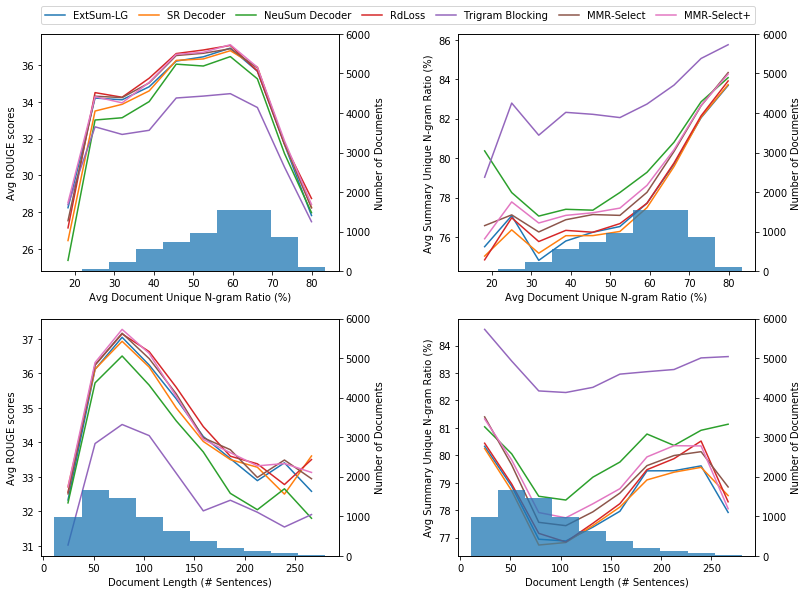}
    \caption{Comparing the average ROUGE scores and average unique n-gram ratios of different models on the Pubmed dataset, conditioned on different degrees of redundancy and lengths of the document (extremely long documents - i.e., 1\% of the dataset are not shown because of space constraints).\footnotemark}
    \label{fig:diff_conditions}
\end{figure*}
\vspace{-1mm}
\subsection{More Insights of the Experiments}
\begin{figure}
    \centering
    \includegraphics[width=0.8\linewidth]{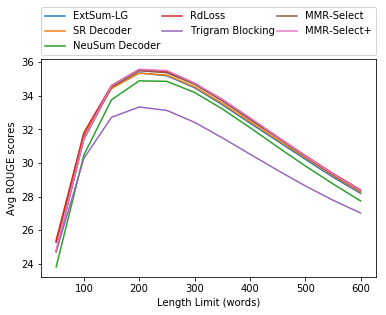}
    \includegraphics[width=0.8\linewidth]{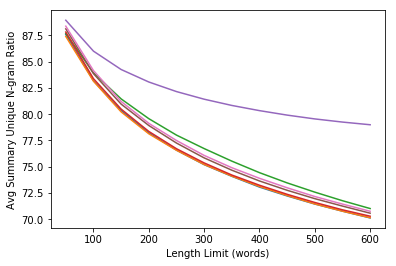}
    \caption{Comparing the average ROUGE scores and average unique n-gram ratios of different models with different word length limits on the Pubmed dataset. See Appendices for similar results on arXiv.}
    \label{fig:diff_length}
\end{figure}
In addition to the main experiment results discussed above, we further explore the performance on informativeness (ROUGE score) and redundancy (Unique N-gram Ratio) of different redundancy reduction methods under two different conditions, namely the degree of redundancy and the length of the source documents. Figure \ref{fig:diff_conditions} shows the results on the Pubmed dataset, while further results of a similar analysis on the arXiv dataset can be found in the Appendices. 
With respect to the degree of redundancy, (upper part of Figure \ref{fig:diff_conditions}), the less redundant the document is, the less impact the redundancy reduction methods have. Among all the methods, although Trigram Blocking works the best with respect to reducing redundancy, it hurts the informativeness the most. However, it is still a good choice for a rather less redundant document (e.g. the documents in the last two bins with avg Unique N-gram Ratio over $0.7$), which is also consistent with the previous works showing the Trigram Blocking works well on the news datasets, which tends to be less redundant (see \S\ref{sec-measure-red}). As for all the other methods, although they have the same trends, MMR-Select+ performs the best on both informativeness and redundancy reduction, especially for the more redundant documents. 

Regarding to the length of the source document (bottom part of Figure \ref{fig:diff_conditions}) 
, as the document become longer, both informativeness and redundancy in the summary generated by all methods increases and then decrease once hitting the peak. MMR-Select+ and MMR-Select are the best choices to balance between the informativeness and redundancy - they are the only two methods having the higher ROUGE scores and higher Unique N-gram ratios across different lengths, even for the short documents with less than 50 sentences.

Besides, we also conduct experiments on generating summaries with different length limit, where we found that our new methods are stable across different summary lengths (Figure. \ref{fig:diff_length}).

\section{Conclusion and Future work}
Balancing sentence importance and redundancy is a key problem in extractive summarization. By examining large summarization datasets, we find that longer documents tend to be more redundant. Therefore in this paper, we systematically explore and compare existing and newly proposed methods  for redundancy reduction in summarizing long documents. 
Experiments indicate that our novel methods achieve SOTA on the ROUGE scores, while significantly reducing  redundancy on two scientific paper datasets (Pubmed and arXiv). Interestingly, we show that redundancy reduction in 
sentence selection 
is more effective than in the sentence scoring phase, a finding to be further investigated . 


Additional venues for future work include experimenting with generating summaries at finer granularity than sentences, as suggested by our analysis of the $\lambda$ parameter. We also intend to explore other ways to assess  redundancy, moving from computing the cosine similarity between sentence embeddings, to a pre-trained neural model for sentence similarity. Finally, 
we plan to run human evaluations to assess the quality of the generated summaries. This is quite challenging for scientific papers, as it requires participants to possess sophisticated domain-specific background knowledge.

\section*{Acknowledgments}
\vspace{-1mm}
We thank 
reviewers and the UBC-NLP group for their insightful comments.
This research was supported by the Language \& Speech Innovation Lab of Cloud BU, Huawei Technologies Co., Ltd.
\bibliographystyle{acl_natbib}
\bibliography{ aacl-ijcnlp2020,anthology}

\begin{thebibliography}{26}
\expandafter\ifx\csname natexlab\endcsname\relax\def\natexlab#1{#1}\fi

\bibitem[{Bi et~al.(2020)Bi, Jha, Croft, and Celikyilmaz}]{Bi2020}
Keping Bi, Rahul Jha, W.~Bruce Croft, and Asli Celikyilmaz. 2020.
\newblock \href {http://arxiv.org/abs/2004.06176} {{AREDSUM: Adaptive
  Redundancy-Aware Iterative Sentence Ranking for Extractive Document
  Summarization}}.

\bibitem[{Carbonell and Goldstein(1998)}]{mmr}
Jaime Carbonell and Jade Goldstein. 1998.
\newblock \href {https://doi.org/10.1145/290941.291025} {The use of mmr,
  diversity-based reranking for reordering documents and producing summaries}.
\newblock In \emph{Proceedings of the 21st Annual International ACM SIGIR
  Conference on Research and Development in Information Retrieval}, SIGIR '98,
  pages 335--336, New York, NY, USA. ACM.

\bibitem[{Cheng and Lapata(2016)}]{cheng-lapata-2016-neural}
Jianpeng Cheng and Mirella Lapata. 2016.
\newblock \href {https://doi.org/10.18653/v1/P16-1046} {Neural summarization by
  extracting sentences and words}.
\newblock In \emph{Proceedings of the 54th Annual Meeting of the Association
  for Computational Linguistics (Volume 1: Long Papers)}, pages 484--494,
  Berlin, Germany. Association for Computational Linguistics.

\bibitem[{Cohan et~al.(2018)Cohan, Dernoncourt, Kim, Bui, Kim, Chang, and
  Goharian}]{discourse-aware}
Arman Cohan, Franck Dernoncourt, Doo~Soon Kim, Trung Bui, Seokhwan Kim, Walter
  Chang, and Nazli Goharian. 2018.
\newblock \href {http://arxiv.org/abs/1804.05685} {A discourse-aware attention
  model for abstractive summarization of long documents}.
\newblock \emph{CoRR}, abs/1804.05685.

\bibitem[{Feigenblat et~al.(2017)Feigenblat, Roitman, Boni, and
  Konopnicki}]{nid_score}
Guy Feigenblat, Haggai Roitman, Odellia Boni, and David Konopnicki. 2017.
\newblock \href {https://doi.org/10.1145/3077136.3080690} {Unsupervised
  query-focused multi-document summarization using the cross entropy method}.
\newblock In \emph{Proceedings of the 40th International ACM SIGIR Conference
  on Research and Development in Information Retrieval}, SIGIR ’17, page
  961–964. Association for Computing Machinery.

\bibitem[{Jung et~al.(2019)Jung, Kang, Mentch, and Hovy}]{Jung2019}
Taehee Jung, Dongyeop Kang, Lucas Mentch, and Eduard Hovy. 2019.
\newblock \href {https://doi.org/10.18653/v1/d19-1327} {{Earlier Isn't Always
  Better: Sub-aspect Analysis on Corpus and System Biases in Summarization}}.
\newblock pages 3322--3333.

\bibitem[{Kedzie et~al.(2018)Kedzie, McKeown, and
  Daum{\'e}~III}]{kedzie-etal-2018-content}
Chris Kedzie, Kathleen McKeown, and Hal Daum{\'e}~III. 2018.
\newblock \href {https://doi.org/10.18653/v1/D18-1208} {Content selection in
  deep learning models of summarization}.
\newblock In \emph{Proceedings of the 2018 Conference on Empirical Methods in
  Natural Language Processing}, pages 1818--1828, Brussels, Belgium.
  Association for Computational Linguistics.

\bibitem[{Lebanoff et~al.(2019)Lebanoff, Muchovej, Dernoncourt, Kim, Kim,
  Chang, and Liu}]{abstractive_analysing}
Logan Lebanoff, John Muchovej, Franck Dernoncourt, Doo~Soon Kim, Seokhwan Kim,
  Walter Chang, and Fei Liu. 2019.
\newblock \href {https://doi.org/10.18653/v1/D19-5413} {Analyzing sentence
  fusion in abstractive summarization}.
\newblock In \emph{Proceedings of the 2nd Workshop on New Frontiers in
  Summarization}, pages 104--110, Hong Kong, China. Association for
  Computational Linguistics.

\bibitem[{Li et~al.(2019)Li, Lei, Qin, and Wang}]{bertrl}
Siyao Li, Deren Lei, Pengda Qin, and William~Yang Wang. 2019.
\newblock \href {https://doi.org/10.18653/v1/D19-1623} {Deep reinforcement
  learning with distributional semantic rewards for abstractive summarization}.
\newblock In \emph{Proceedings of the 2019 Conference on Empirical Methods in
  Natural Language Processing and the 9th International Joint Conference on
  Natural Language Processing (EMNLP-IJCNLP)}, pages 6040--6046, Hong Kong,
  China. Association for Computational Linguistics.

\bibitem[{Lin(2004)}]{lin-2004-rouge}
Chin-Yew Lin. 2004.
\newblock \href {https://www.aclweb.org/anthology/W04-1013} {{ROUGE}: A package
  for automatic evaluation of summaries}.
\newblock In \emph{Text Summarization Branches Out}, pages 74--81, Barcelona,
  Spain. Association for Computational Linguistics.

\bibitem[{Lin et~al.(2009)Lin, Bilmes, and Xie}]{Lin2009}
Hui Lin, Jeff Bilmes, and Shasha Xie. 2009.
\newblock \href {https://doi.org/10.1109/ASRU.2009.5373486} {{Graph-based
  submodular selection for extractive summarization}}.
\newblock \emph{Proceedings of the 2009 IEEE Workshop on Automatic Speech
  Recognition and Understanding, ASRU 2009}, pages 381--386.

\bibitem[{Liu and Lapata(2019)}]{liu-lapata-2019-text}
Yang Liu and Mirella Lapata. 2019.
\newblock \href {https://doi.org/10.18653/v1/D19-1387} {Text summarization with
  pretrained encoders}.
\newblock In \emph{Proceedings of the 2019 Conference on Empirical Methods in
  Natural Language Processing and the 9th International Joint Conference on
  Natural Language Processing (EMNLP-IJCNLP)}, pages 3730--3740, Hong Kong,
  China. Association for Computational Linguistics.

\bibitem[{Lloret and Sanz(2013)}]{Tackling_redundancy}
Elena Lloret and Manuel Sanz. 2013.
\newblock \href {https://doi.org/10.1016/j.csi.2012.08.001} {Tackling
  redundancy in text summarization through different levels of language
  analysis}.
\newblock \emph{Computer Standards \& Interfaces}, 35:507–518.

\bibitem[{Nallapati et~al.(2016{\natexlab{a}})Nallapati, Zhai, and
  Zhou}]{summarunner}
Ramesh Nallapati, Feifei Zhai, and Bowen Zhou. 2016{\natexlab{a}}.
\newblock \href {http://arxiv.org/abs/1611.04230} {Summarunner: {A} recurrent
  neural network based sequence model for extractive summarization of
  documents}.
\newblock \emph{CoRR}, abs/1611.04230.

\bibitem[{Nallapati et~al.(2016{\natexlab{b}})Nallapati, Zhou, dos Santos,
  Gul{\c{c}}ehre, and Xiang}]{nallapati-etal-2016-abstractive}
Ramesh Nallapati, Bowen Zhou, Cicero dos Santos, {\c{C}}a{\u{g}}lar
  Gul{\c{c}}ehre, and Bing Xiang. 2016{\natexlab{b}}.
\newblock \href {https://doi.org/10.18653/v1/K16-1028} {Abstractive text
  summarization using sequence-to-sequence {RNN}s and beyond}.
\newblock In \emph{Proceedings of The 20th {SIGNLL} Conference on Computational
  Natural Language Learning}, pages 280--290, Berlin, Germany. Association for
  Computational Linguistics.

\bibitem[{Narayan et~al.(2018)Narayan, Cohen, and
  Lapata}]{narayan-etal-2018-dont}
Shashi Narayan, Shay~B. Cohen, and Mirella Lapata. 2018.
\newblock \href {https://doi.org/10.18653/v1/D18-1206} {Don{'}t give me the
  details, just the summary! topic-aware convolutional neural networks for
  extreme summarization}.
\newblock In \emph{Proceedings of the 2018 Conference on Empirical Methods in
  Natural Language Processing}, pages 1797--1807, Brussels, Belgium.
  Association for Computational Linguistics.

\bibitem[{Paulus et~al.(2017)Paulus, Xiong, and Socher}]{rl-abstractive}
Romain Paulus, Caiming Xiong, and Richard Socher. 2017.
\newblock \href {http://arxiv.org/abs/1705.04304} {A deep reinforced model for
  abstractive summarization}.
\newblock \emph{CoRR}, abs/1705.04304.

\bibitem[{Pennington et~al.(2014)Pennington, Socher, and Manning}]{glove}
Jeffrey Pennington, Richard Socher, and Christopher~D. Manning. 2014.
\newblock \href {http://www.aclweb.org/anthology/D14-1162} {Glove: Global
  vectors for word representation}.
\newblock In \emph{Empirical Methods in Natural Language Processing (EMNLP)},
  pages 1532--1543.

\bibitem[{Peyrard et~al.(2017)Peyrard, Botschen, and
  Gurevych}]{peyrard-etal-2017-learning}
Maxime Peyrard, Teresa Botschen, and Iryna Gurevych. 2017.
\newblock \href {https://doi.org/10.18653/v1/W17-4510} {Learning to score
  system summaries for better content selection evaluation.}
\newblock In \emph{Proceedings of the Workshop on New Frontiers in
  Summarization}, pages 74--84, Copenhagen, Denmark. Association for
  Computational Linguistics.

\bibitem[{Qian et~al.(2019)Qian, Muaz, Zhang, and
  Hyun}]{qian-etal-2019-reducing}
Yusu Qian, Urwa Muaz, Ben Zhang, and Jae~Won Hyun. 2019.
\newblock \href {https://doi.org/10.18653/v1/P19-2031} {Reducing gender bias in
  word-level language models with a gender-equalizing loss function}.
\newblock In \emph{Proceedings of the 57th Annual Meeting of the Association
  for Computational Linguistics: Student Research Workshop}, pages 223--228,
  Florence, Italy. Association for Computational Linguistics.

\bibitem[{Ren et~al.(2016)Ren, Wei, Chen, Ma, and Zhou}]{redundancy_aware}
Pengjie Ren, Furu Wei, Zhumin Chen, Jun Ma, and Ming Zhou. 2016.
\newblock \href {https://www.aclweb.org/anthology/C16-1004} {A redundancy-aware
  sentence regression framework for extractive summarization}.
\newblock In \emph{Proceedings of {COLING} 2016, the 26th International
  Conference on Computational Linguistics: Technical Papers}, pages 33--43,
  Osaka, Japan. The COLING 2016 Organizing Committee.

\bibitem[{Saggion and Poibeau(2013)}]{summary_eval}
Horacio Saggion and Thierry Poibeau. 2013.
\newblock \href {https://doi.org/10.1007/978-3-642-28569-1_1} {\emph{Automatic
  Text Summarization: Past, Present and Future}}, pages 3--21. Springer Berlin
  Heidelberg, Berlin, Heidelberg.

\bibitem[{Stewart and Carbonell(1998)}]{longdoc_more_redundancy}
Jade Stewart and Jaime Carbonell. 1998.
\newblock \href {https://doi.org/10.3115/1119089.1119120} {Summarization: (1)
  using mmr for diversity - based reranking and (2) evaluating summaries}.
\newblock pages 181--195.

\bibitem[{Vinyals et~al.(2015)Vinyals, Fortunato, and Jaitly}]{pointer_network}
Oriol Vinyals, Meire Fortunato, and Navdeep Jaitly. 2015.
\newblock \href {http://papers.nips.cc/paper/5866-pointer-networks.pdf}
  {Pointer networks}.
\newblock In C.~Cortes, N.~D. Lawrence, D.~D. Lee, M.~Sugiyama, and R.~Garnett,
  editors, \emph{Advances in Neural Information Processing Systems 28}, pages
  2692--2700. Curran Associates, Inc.

\bibitem[{Xiao and Carenini(2019)}]{xiao-carenini-2019-extractive}
Wen Xiao and Giuseppe Carenini. 2019.
\newblock \href {https://doi.org/10.18653/v1/D19-1298} {Extractive
  summarization of long documents by combining global and local context}.
\newblock In \emph{Proceedings of the 2019 Conference on Empirical Methods in
  Natural Language Processing and the 9th International Joint Conference on
  Natural Language Processing (EMNLP-IJCNLP)}, pages 3002--3012, Hong Kong,
  China. Association for Computational Linguistics.

\bibitem[{Zhou et~al.(2018)Zhou, Yang, Wei, Huang, Zhou, and Zhao}]{NeuSum}
Qingyu Zhou, Nan Yang, Furu Wei, Shaohan Huang, Ming Zhou, and Tiejun Zhao.
  2018.
\newblock \href {https://doi.org/10.18653/v1/P18-1061} {Neural document
  summarization by jointly learning to score and select sentences}.
\newblock In \emph{Proceedings of the 56th Annual Meeting of the Association
  for Computational Linguistics (Volume 1: Long Papers)}, pages 654--663,
  Melbourne, Australia. Association for Computational Linguistics.

\end{thebibliography}
\appendix
\section{Appendices}
In these Appendices, we show more analysis of the experimental results. 
\subsection{Analysis on arXiv Dataset under conditions}
\begin{figure*}[htb!]
    \centering
    \includegraphics[width=0.8\linewidth]{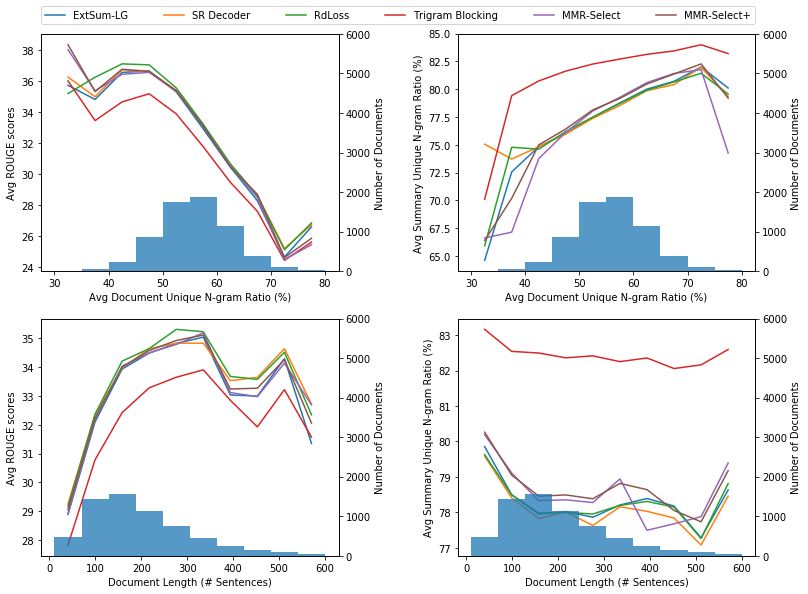}
    \caption{Comparing the average ROUGE scores and average unique n-gram ratios of different models on the arXiv dataset, conditioned on different degrees of redundancy and lengths of the document.\footnotemark}
    \label{fig:analysis_arxiv}
\end{figure*}
Figure \ref{fig:analysis_arxiv} shows the performance on informativeness (ROUGE score) and redundancy (Unique N-gram Ratio) of different redundancy reduction methods under different conditions on the arXiv dataset. Comparing with the Pubmed dataset, the documents in the arXiv dataset tend to be 
longer and more redundant, as the majority of the documents in the Pubmed dataset have less than 100 sentences with average Unique N-gram Ratio in the $0.5-0.6$ interval, while the majority of the documents in the arXiv dataset have number of sentences in the range 100 to 300 with average Unique N-gram Ratio in the $0.6-0.7$ interval. Consistent with the result on the Pubmed dataset, the Trigram Blocking method is the best choice for rather less redundant documents (with average Unique N-gram Ratio larger than $0.7$), and the MMR-Select+ is the one better or equivalent to the original model across different degree of redundancy, ignoring the outliers. With respect to the length of the documents, the MMR-Select+ and MMR-Select are consistently the most effective methods for balancing redundancy and informativeness on documents with different length.

\subsection{ Analysis on Selection Overlap}
\begin{table*}[]
    \centering
    \resizebox{\linewidth}{!}{\begin{tabular}{c|c|c|c|c|c|c|c}
      -  &  ExtSumLG &+SR &+NeuSum &+RdLoss &+Tri-Block&+MMR-Select &+MMR-Select+ \\
      \hline
      ExtSumLG &  100.00&72.84&52.00&77.70&60.77&87.71&85.75\\
      \quad+SR  & 72.66&100.00&49.73&70.29&52.24&69.78&70.64\\
      \quad+Neusum & 60.44&57.94&100.00&60.77&48.47&60.38&61.07\\
      \quad+RdLoss & 80.84&73.32&54.40&100.00&57.67&79.03&80.08 \\
      \quad+Tri-Block   & 64.85&55.89&44.51&59.15&100.00&64.72&64.38\\
      \quad+MMR-Select   & 90.49&72.17&53.59&78.37&62.56&100.00&91.15\\
      \quad+MMR-Select+   &88.66&73.22&54.33&79.58&62.38&91.35&100.00\\
      \hhline{========}
      \# Sent. Sel.&36979&36888&42981&38476&39463&38151&38236\\
      \# words/Sent &40.66 &40.84 &33.38 &38.95 &37.21 &39.35 &39.31\\
    \end{tabular}}
    \caption{Micro overlap ratio (\%) between the selections of different methods and the total number and the average length of selected sentences in the test set of Pubmed.}
    \label{tab:overlap_pubmed}
\end{table*}

\begin{table*}[]
    \centering
    \resizebox{\linewidth}{!}{\begin{tabular}{c|c|c|c|c|c|c|c}
       -  &  ExtSumLG &+SR &+NeuSum &+RdLoss &+Tri-Block&+MMR-Select &+MMR-Select+ \\
       \hline
      ExtSumLG &  100.00&72.06&-&76.51&56.22&75.04&80.21\\
      \quad+SR  & 73.84&100.00&-&69.07&49.16&62.82&67.03\\
      \quad+Neusum & -&-&-&-&-&-&-\\
      \quad+RdLoss &  79.88&70.38&-&100.00&53.00&67.81&72.57\\
      \quad+Tri-Block   & 64.59&55.12&-&58.33&100.00&60.34&62.55\\
      \quad+MMR-Select   & 88.93&72.65&-&76.97&62.23&100.00&93.13\\
      \quad+MMR-Select+   &89.96&73.36&-&77.96&61.06&88.14&100.00\\
      \hhline{========}
      \# Sent. Sel.&39698&40681&-&41448&45611&47045&44526\\
      \# words/Sent &36.26&35.52&-&34.50&30.86&30.73&32.40\\
    \end{tabular}}
    \caption{Micro overlap ratio (\%) between the selections of different methods and the total number and the average length of selected sentences in the test set of arXiv.}
    \label{tab:overlap_arxiv}
\end{table*}
To explore the difference made by applying different redundancy reduction methods on the original method(ExtSumLG), we compare the selected sentences by all the methods, and show the overlap ratios between every two methods, as well as the total number and the average length of selected sentences in the test set, in Table \ref{tab:overlap_pubmed} and Table \ref{tab:overlap_arxiv} for Pubmed dataset and arXiv dataset respectively. As we can see from the tables, except for the SR Decoder, all the other methods tend to select more and shorter sentences than the original summarizer. Regarding the overlap between the original method and the others, we observe that among all the three categories, the methods in category A tend to produce large differences, since these methods change the structure of the original model. Comparing the methods in Category C, around $36\%$ of the sentences are regarded as redundant by Trigram Blocking, which means $36\%$ of the sentences have trigram-overlap with other selected sentences, while only around $10\%$ sentences are regarded as redundant by MMR-Select. As the ROUGE scores of MMR-Select are much better than Trigram Blocking on both datasets, this is in line with our analysis in Section \ref{result_and_analysis}, Triagram Blocking dropping some important sentences incorrectly. Interestingly,
we notice that the overlap ratio between Trigram Block and MMR-Select is considerably larger than the overlap ratio of Trigram Block with original method (ExtSumLG) on both datasets. This indicates that there are some sentences, not selected by the original method, which are considered to be important by both the Trigram Blocking and MMR-Select methods.

\subsection{Analysis on Recall and Precision of ROUGE Scores }
 \begin{table*}[h!]
    \centering
    \resizebox{0.85\linewidth}{!}{
    \begin{tabular}{c|l|c|c|c|c|c|c}
    \hline
    \multirow{2}{*}{Categ.}&\multirow{2}{*}{Model}&\multicolumn{6}{c}{Pubmed}\\
    \cline{3-8}
      &   &\multicolumn{2}{c|}{ROUGE-1}&\multicolumn{2}{c|}{ROUGE-2}&\multicolumn{2}{c}{ROUGE-L} \\
    \cline{3-8}
      &   &Prec.&Recall&Prec.&Recall&Prec.&Recall\\  
    \hline
C&       Naive MMR&36.45 &42.56 &11.05 &12.64 &31.39 &36.53 \\
-&       ExtSum-LG\footnotemark   &44.05&51.08  &19.82&22.71 &39.74&45.97 \\
\hline
A&       \quad+SR Decoder  &\textcolor{Red}{44.00}&\textcolor{Green}{51.10}  &\textcolor{Red}{19.75} &\textcolor{Red}{22.68} &\textcolor{Red}{39.66}&\textcolor{Red}{45.96} \\
A&       \quad+NeuSum Decoder &\textbf{\textcolor{Green}{44.36}}&\textcolor{Red}{49.24}  &\textcolor{Red}{19.74}&\textcolor{Red}{21.58}  &\textbf{\textcolor{Green}{40.29}} &\textcolor{Red}{44.62} \\
B&        \quad\textbf{+RdLoss}&\textcolor{Green}{44.30}&\textcolor{Green}{51.09}  &\textbf{\textcolor{Green}{20.11}}&\textcolor{Green}{22.88}  &\textcolor{Green}{40.09}&\textcolor{Green}{46.11}\\       
C&       \quad+Trigram Blocking &\textcolor{Red}{42.67}&\textcolor{Red}{48.54}  &\textcolor{Red}{17.51}&\textcolor{Red}{19.73}  &\textcolor{Red}{38.45}&\textcolor{Red}{43.64}\\
C&       \quad\textbf{+MMR-Select}&\textcolor{Green}{44.25}&\textcolor{Green}{51.09}  &\textcolor{Green}{19.98} &\textcolor{Green}{22.75} &\textcolor{Green}{40.08}&\textcolor{Green}{46.07}\\
C&        \quad\textbf{+MMR-Select+} &\textcolor{Green}{44.28}&\textbf{\textcolor{Green}{51.27}}  &\textcolor{Green}{20.01}&\textbf{\textcolor{Green}{22.86}}  &\textcolor{Green}{40.03}&\textbf{\textcolor{Green}{46.24}}\\
    \hline
    \end{tabular}}
    \caption{Rouge Recall and Precision of different summarization models on the Pubmed dataset.  \textcolor{Green}{Green} numbers means it's \textbf{better} than ExtSum-LG on the certain metric, and the \textcolor{Red}{red} numbers means worse.}
    \label{tab:precision_recall_pubmed}
\end{table*}

 \begin{table*}[h!]
    \centering
    \begin{tabular}{c|l|c|c|c|c|c|c}
    \hline
    \multirow{2}{*}{Categ.}&\multirow{2}{*}{Model}&\multicolumn{6}{c}{Arxiv}\\
    \cline{3-8}
      &   &\multicolumn{2}{c|}{ROUGE-1}&\multicolumn{2}{c|}{ROUGE-2}&\multicolumn{2}{c}{ROUGE-L} \\
    \cline{3-8}
      &   &Prec.&Recall&Prec.&Recall&Prec.&Recall\\  
    \hline
C&       Naive MMR&29.61 &42.69 &7.45 &10.78 &24.92 &35.82 \\
-&       ExtSum-LG\footnotemark   &38.60&54.64 &15.38&22.00  & 34.17& 48.26\\
\hline
A&       \quad+SR Decoder &\textcolor{Green}{38.65} &\textbf{\textcolor{Green}{54.99}}  &\textcolor{Green}{15.47}&\textcolor{Green}{22.28}  &\textcolor{Green}{34.24} &\textbf{\textcolor{Green}{48.64}}\\
A&       \quad+NeuSum Decoder &-&-&-&-&-&- \\
B&        \quad\textbf{+RdLoss} &\textbf{\textcolor{Green}{38.92}}&\textcolor{Green}{54.77}  &\textbf{\textcolor{Green}{15.68}}&\textbf{\textcolor{Green}{22.29}}  &\textbf{\textcolor{Green}{34.60}}&\textcolor{Green}{48.59}\\       
C&       \quad+Trigram Blocking&\textcolor{Red}{38.04} & \textcolor{Red}{52.71}&\textcolor{Red}{13.98} &\textcolor{Red}{19.47} &\textcolor{Red}{33.71} &\textcolor{Red}{46.61}\\
C&       \quad\textbf{+MMR-Select}&\textcolor{Green}{38.85} &\textcolor{Red}{54.33} &\textcolor{Green}{15.39} &\textcolor{Red}{21.74} &\textcolor{Green}{34.56} &\textcolor{Red}{48.24}\\
C&        \quad\textbf{+MMR-Select+} &\textcolor{Green}{38.75} &\textcolor{Green}{54.67} &\textcolor{Green}{15.41} &\textcolor{Red}{21.96} &\textcolor{Green}{34.44} &\textcolor{Green}{48.51}\\
    \hline
    \end{tabular}
    \caption{Rouge Recall and Precision of different summarization models on the Pubmed dataset.  \textcolor{Green}{Green} numbers means it's \textbf{better} than ExtSum-LG on the certain metric, and the \textcolor{Red}{red} numbers means worse.}
    \label{tab:precision_recall_arxiv}
\end{table*}
We also provide the Precision and Recall of the ROUGE scores in the main experiment, the results of Pubmed and arXiv datasets are shown in Table \ref{tab:precision_recall_pubmed} and Table \ref{tab:precision_recall_arxiv}, respectively. It is interesting to see that the NeuSum Decoder tends to have a high precision but low recall, indicating that the generated summaries tend to be shorter and contain less useful information than the original method. 
\subsection{Analysis on the Relative Position of Selections}
\begin{figure*}[h!]
    \centering
    \includegraphics[width=0.4\linewidth]{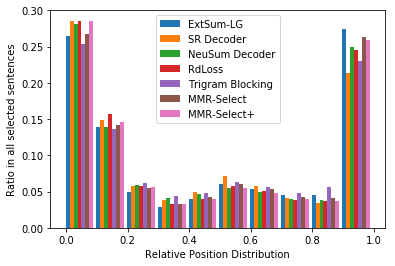}
    \includegraphics[width=0.4\linewidth]{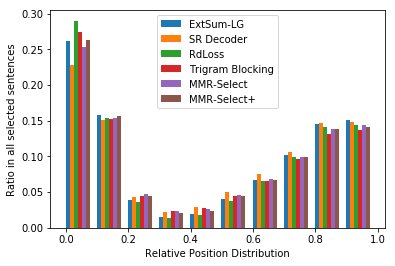}
    \caption{The relative position distribution of different redundancy reduction methods on Pubmed(left) and arXiv(right) datasets.}
    \label{fig:relative_position}
\end{figure*}
We also show the relative position distribution of the selected sentences on both datasets in Figure \ref{fig:relative_position} to verify if any redundancy reduction method has a particular  tendency to select sentences in particular position of the documents. However, as shown in the figure, the trends are all rather similar for all methods.
\end{document}